\documentclass{article} 
\usepackage{iclr2025_conference,times}

\usepackage{amsmath,amsfonts,bm}

\def\eqref#1{equation~\ref{#1}}

\def\1{\bm{1}}

\DeclareMathAlphabet{\mathsfit}{\encodingdefault}{\sfdefault}{m}{sl}
\SetMathAlphabet{\mathsfit}{bold}{\encodingdefault}{\sfdefault}{bx}{n}

\usepackage[hidelinks,colorlinks]{hyperref}

\usepackage{algorithm}
\usepackage{algpseudocode}

\usepackage{newfloat}
\usepackage{listings}
\usepackage{amssymb}
\usepackage{comment}
\usepackage{makecell}
\usepackage{xcolor}
\usepackage{multicol}
\usepackage{multirow}
\usepackage{booktabs}
\usepackage{url}
\usepackage{graphicx}

\makeatletter
\def\blfootnote{\xdef\@thefnmark{}\@footnotetext}
\makeatother

\title{DriveScape: Towards High-Resolution Controllable Multi-View Driving Video Generation}

\author{Wei Wu$^{1,2}$,
Xi Guo$^{2}$,
Weixuan Tang$^{2}$,
Tingxuan Huang$^{3}$, \\
\textbf{Chiyu Wang}$^{2}$,
\textbf{Dongyue Chen}$^{3}$,
\textbf{Chenjing Ding}$^{2\dag}$\\
$^{1}$Tsinghua University
\enspace
$^{2}$Sensetime Research
\enspace
$^{3}$Northeastern University\\
\texttt{\{wuwei,guoxi,dingchenjing\}@sensetime.com},\\
\texttt{\{tangweixuan, wangchiyu\}@senseauto.com},\enspace\\
\texttt{\{huangtingxuan,chendongyue\}@stu.neu.edu.cn} \\
}

\iclrfinalcopy 
\begin{document}

\maketitle

\begin{abstract}
Recent advancements in generative models have provided promising solutions for synthesizing realistic driving videos, which are crucial for training autonomous driving perception models. However, existing approaches often struggle with multi-view video generation due to the challenges of integrating 3D information while maintaining spatial-temporal consistency and effectively learning from a unified model.
We propose DriveScape, an end-to-end framework for multi-view, 3D condition-guided video generation, capable of producing 1024 x 576 high-resolution videos at 10Hz. Unlike other methods limited to 2Hz due to the 3D box annotation frame rate, DriveScape overcomes this with its ability to operate under sparse conditions. Our Bi-Directional Modulated Transformer (BiMot) ensures precise alignment of 3D structural information, maintaining spatial-temporal consistency. DriveScape excels in video generation performance, achieving state-of-the-art results on the nuScenes dataset with an FID score of 8.34 and an FVD score of 76.39. Code will be available at \href{https://metadrivescape.github.io/papers_project/drivescapev1/index.html}{our project homepage}.
\end{abstract}

\blfootnote{
$^{\dag}$Corresponding authors.
Work in progress.
}

\section{Introduction}
Autonomous driving has gained significant attention, highlighting the need for precise environmental understanding to ensure safe driving decisions (\cite{uniad, genad}). Bird's-eye view (BEV) maps, generated from multi-view images, serve as a crucial structured representation for tasks such as 3D object detection, segmentation, tracking, depth estimation, and trajectory prediction (\cite{bevformer, bevformerv2, bevdet, bevdet4d, cvt, bevtrack, bevdepth, liao2024cognitive}). However, acquiring high-quality multi-view video data is challenging due to the high labeling costs, driving the demand for generating temporally consistent video data aligned with real-world distributions and 3D labels. Recent generative models have shown the potential of synthetic data to enhance visual tasks (\cite{magicdrive, bevgen, bevcontrol}), while diffusion models have advanced the generation of diverse and realistic driving videos, crucial for building autonomous driving systems.

\definecolor{myLightBlue}{rgb}{0.3, 0.3, 1.0}
\definecolor{myLightRed}{rgb}{1.0, 0.3, 0.3}
\begin{table}[ht]
\centering
\resizebox{0.9\linewidth}{!}{
\begin{tabular}{c|c|c|c}
\hline                   
\textbf{Model} &  \makecell{Spatial Res. } & \makecell{Temporal Res. (Hz)} & \makecell{sparse condition}  \\ \hline
Drive-WM & 192 $\times$ 384 & 2 & $\times$   \\
WoVoGen  & 256 $\times$ 448 & 2 & $\times$     \\ 
Delphi   & 512 $\times$ 512 & 2 & $\times$       \\ 
GenAD    & 256 $\times$ 448  & 2  & $\times$     \\ 
MagicDrive  & 272 $\times$ 736  & - & $\times$      \\ 
DriveDreamer 1\&2 & 448 $\times$ 256 & 2 & $\times$   \\ 
DriveDiffusion  & 512 $\times$ 512 & 2 & $\times$  \\ 
Panacea & 256 $\times$ 512 & 2 & $\times$ \\ \hline
Ours (DriveScape) & \textbf{\textcolor{myLightRed}{576 $\times$ 1024}} & \textbf{\textcolor{myLightRed}{2 - 10}} & $\checkmark$ \\ \hline
\end{tabular}
}
\caption{Real-world multi-view driving controllable world models. DriveScape (ours) operates at a high spatiotemporal resolution, offering support for sparse layout controllability, as well as unified control for both mono-condition and multi-condition scenarios.}
\label{tab:worldmodel compare}
\vspace{-0.2cm}
\end{table}

The most relevant approach to us is the \textbf{generation of multi-view driving videos controlled by 3D conditions}. \textbf{Firstly}, as shown in Tab \ref{tab:worldmodel compare}, most methods generate images at low spatial and temporal resolutions. Since they do not support sparse conditions, they cannot generate high-frequency videos from lower-frequency inputs, which is an important capability for autonomous driving, where annotations are far more costly than video capturing. \textbf{Secondly}, most methods lack fine-grained layout controllability, which is essential for flexible condition adjustment across various perception tasks, ultimately reducing generation costs, as obtaining all types of layouts can be expensive. \textbf{Thirdly}, other approaches (\cite{magicdrive, drivediffusion, drivedreamer}) directly concatenate multi-type control information without considering the synergy and alignment of each structural component in both spatial and temporal dimensions, leading to inconsistencies and degraded results. \textbf{Finally}, due to the complexity of multi-view driving video generation, certain approaches (\cite{drivediffusion, drivedreamer}) involve a multi-stage scheme relying on complex pipelines. DrivingDiffusion (\cite{drivediffusion}) uses two separate models and multiple stages to generate frames and extend the video sequentially, requiring post-processing. This complex cascaded approach is inefficient, prone to cumulative errors, and may lack robustness, thus limiting its practical application.

To address these challenges, we propose DriveScape, a solution for multi-view 3D condition-guided video generation. We introduce a streamlined and efficient training and inference pipeline, without any post-processing, that supports high spatial and temporal resolution, enabling sparse condition control to generate multi-view, high-resolution, high-frame-rate videos. We also introduce a Bi-Directional Condition Alignment module to achieve effective alignment and synergy between various 3D road structural information. By integrating multiple conditions in an aligned latent space and incorporating conditions from a broader to a more refined level, we can realize fine-grained layout controllability and enable precise control, significantly enhancing the realism of the generated videos. Furthermore, we fully leverage temporal and spatial information to create a unified model for consistent multi-view video generation.

In summary, our key contributions are as follows:
\begin{itemize}
\item[•] We propose an effective and streamlined framework for multi-view driving video generation without complex post-processing. To the best of our knowledge, \textbf{our model is the first to achieve high-resolution, high-frame-rate, sparse condition control for multi-view driving video generation}.
\item[•] We introduce a Bi-Directional Condition Alignment module to facilitate diverse modalities of condition guidance and sparse condition control in video generation tasks. Experiments demonstrate that our model achieves decoupled, sparse, and precise control of dynamic foreground and static background. 
\item[•] We achieve state-of-the-art video synthesis performance on the nuScenes dataset, excelling not only in the generation quality of multi-view images and videos but also across various perception applications.
\end{itemize}

\begin{figure*}[htbp]
\centering
\includegraphics[width=1.0\linewidth]{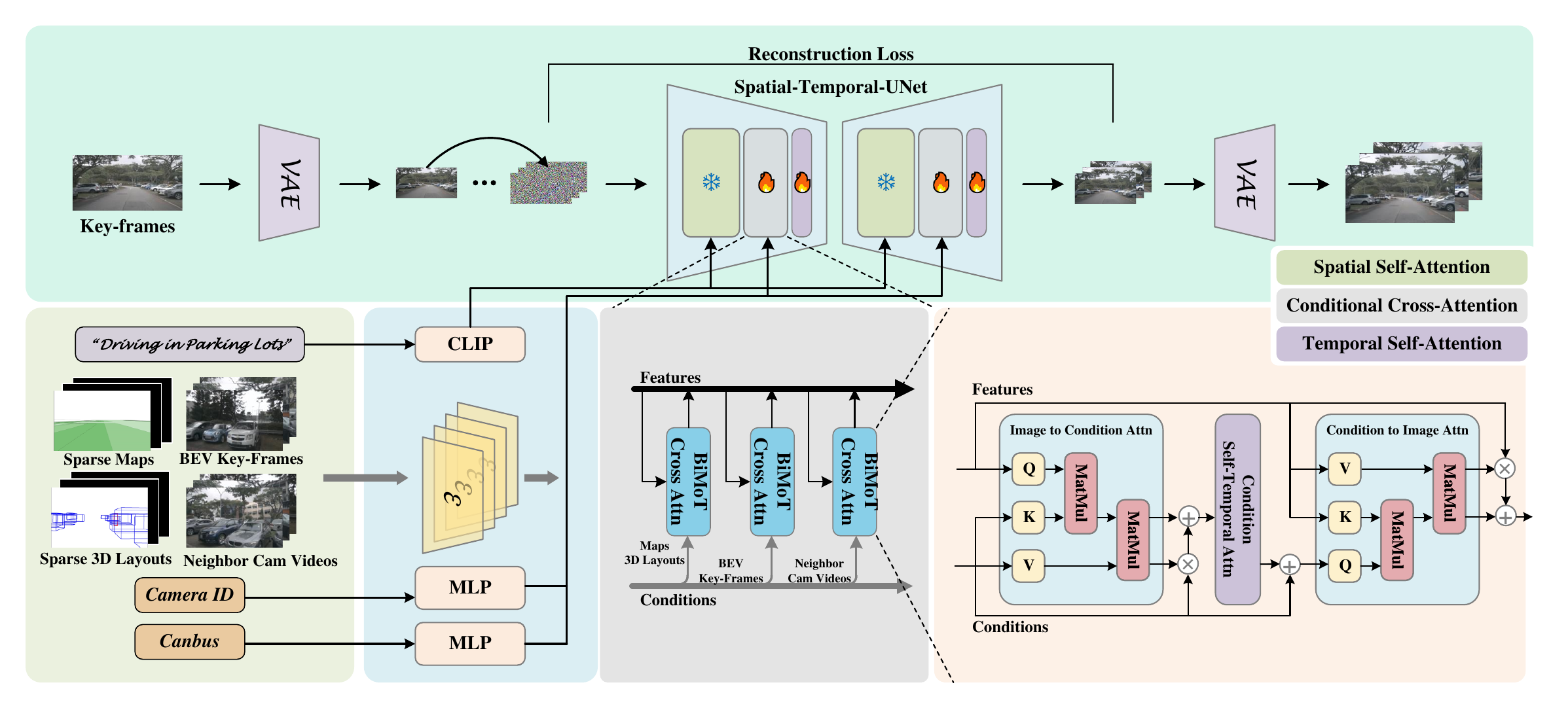}
\caption{End-to-End Multi-view Video Generation pipeline. We use learnable embedding vectors to represent different cameras and categorize camera views as key views and neighbour views. Our training scheme guides the generation of key view videos through their neighboring frames. Additionally, we introduce the key-frames condition along with a training and inference scheme to ensure multi-view consistency simultaneously. Furthermore, our model does not require any post-refine process as DriveDiffusion (\cite{drivediffusion}). It is capable of learning multi-view and temporal consistency simultaneously, resulting in high-fidelity street-view synthesis.}
\label{fig:network}
\end{figure*}

\section{Related Works}
\subsection{Controllable generation}
With the advent of diffusion models, great progress has been made in the field of text-to-video generation (\cite{an2023latent, blattmann2023align, guo2023animatediff, he2022latent, ho2022imagen, blattmann2023stable,singer2022makeavideo, wang2023videofactory, zhou2023magicvideo}). Video LDM (\cite{VideoLDM}) adopts a latent diffusion pipeline where diffusion denoise progresses on the image latent, speeding up the denoising process greatly. Text cannot accurately control video generation, so later methods added image blocks together with text as prompt information to be input into the denoising network for control (\cite{zhang2024moonshot}). What we hope to generate is  highly realistic street videos. This scene is very complex, with many elements and interactions between elements (such as complex street arrangements, moving cars, etc.), so we need more information for fine control, not only just pictures and text. In our method, we combine road maps, 3D bounding boxes and BEV key-frames to control video generation finely.

\subsection{Multi-view video generation}
The multi-view consistency and temporal consistency are two crucial problems in multi-view video generation. To keep multi-view consistency,  MVDiffusion (\cite{tang2023mvdiffusion}) proposed a correspondence-aware attention module to align the information from multiple views. (\cite{tseng2023consistent}) utilise epipolar geometry to regularize the consistency between different views. MagicDrive (\cite{magicdrive}) leverage camera pose, bounding box and road map as prior, then insert the extra cross-view attention block to improve consistency. However, those methods can only generate multi-view images instead of video. 

\subsection{Street view generation}
Most street view generation models condition on 2D layouts, such as BEV maps, 2D bounding boxes and semantic segmentation. BEVGen (\cite{bevgen}) approaches street view generation by encapsulating all semantics within BEV. BEVControl (\cite{bevcontrol}) proposes a two-stage method generating multi-view urban scene images from a BEV layout, in which a controller generates foreground and background objects, and a coordinator combines them together, preserving visual consistency across different views. In addition, projecting 3D information to 2D loses the 3D geometric information. Thus, extending them directly to video generation may result in inconsistencies between multiple frames. We introduce 3D bounding boxes as one of the conditions to guide generation. DrivingDffusion (\cite{drivediffusion}) proposes a multi-stage scheme with two models and 2 stages of post-processing to generate frames and extend the video separately. However, those methods rely on a multi-stage pipeline which is complex, while our method uses an effective and efficient end-to-end pipeline.
 
\section{Method}

\subsection{Overview}
The overview of DriveScape is depicted in Fig \ref{fig:network}. Operating on the LDM (\cite{ldm}) pipeline, DriveScape generates street-view videos conditioned on scene annotations BEV map,  3D bounding boxes, ego state and text for each view. We introduce the unified model in Section \ref{sec:UnifiedModel}. We achieve one unified model for high resolution multi-view video generation without complex post-processing and any post-refine process. In addition, the Bi-Directional Condition Alignment module will be introduced in Section \ref{sec:BiDirectionalModulatedTransformer}, which can achieve effective alignment and synergy between various 3D road structural information and sparse condition control. 

\subsection{Unified Model}
\label{sec:UnifiedModel}

Our unified model is mainly composed of Unet with spatial and temporal convolution and attention. With the first multi-view frames $x_{0:N - 1}^0$ ($N$ denotes the number of views), DriveScape can predict the next frames $x_{0:N - 1}^{1:T}$ ($T$ denotes the number of frames) with various conditions including BEV maps $m_{0:N - 1}^{0:T}$, 3D BBoxs $b_{0:N - 1}^{0:T}$,BEV Key-Frames and Neighbor Camera Videos. BEV maps and 3D BBoxs will be projected to the camera’s First-Person View (FPV) and subsequently encoded using convolutional blocks.

As illustrated in Fig~\ref{fig:network}, we utilize learnable embedding vectors ${e_i}$ to represent distinct cameras. Same as canbus information (velocity and direction angle), these vectors are then input into a multilayer perceptron (MLP) and a Sigmoid Linear Unit (SiLU) function. This process enables the embedding vectors to engage in cross-attention with image latent, resulting in a unified model capable of predicting videos from different perspectives. 

However, embedding camera information can not achieve consistency across spatial and temporal dimensions, since it can only represent the global characteristic of each camera over the whole dataset. Consistency will be ensured by the design of the model and the training scheme. On the one hand, to achieve temporal consistency, we fully leverage temporal information by incorporating temporal attention in layers of the latent diffusion UNet (\cite{unet2015}) and control alignment module which will be discussed in Sec.BiDirectionalModulatedTransformer(BiMOT). On the other hand, we also introduce the KeyFrames condition as well as the training and inference scheme to ensure multi-view consistency simultaneously.

\subsection{Training Stage.}

\noindent \textbf{Generation with nerighbor frames} DriveScape categorizes camera views as key views and neighbor views, as illustrated in  Fig~\ref{fig:network2}. Frames of the same type have minimal visible overlap. In fact, there is almost no discernible overlapping regions between them. As a result, our training scheme will steer the generation of key view videos through their neighboring frames. During the training stage, the neighbor views and key views are sequentially selected. While only in the neighbor views training process, video conditions of its neighbor camera which is the near key views are fed into the network. DriveScape produces videos for the key-views in sequence and stores them for subsequent training phases. Once the key views are prepared, they are involved in cross-attention with image latent to ensure that the neighbor view is guided by its neighbors.
 
\noindent \textbf{Generation with key-frame conditions} One challenge we encounter is that  key-views are generated without specific multi-view constraints input which will lead to degraded performance. In most cases, this is reasonable due to almost no overlapping regions among them. However, in the context of long-term generation, these views are interrelated. For instance, when overtaking occurs, these vehicles are initially observed by the rear camera and subsequently captured by the front camera after a certain duration. Therefore, the key-frame conditions are proposed which are the start frame of all views to correlate with image latent. Combined with the temporal attention, the model can learn the correlation of all views in the beginning and the preceding frames of all neighbor cameras. 

\noindent \textbf{Generation with sparse conditions} Our model supports training with videos at various frame rates, ranging from 2Hz to 10Hz, while maintaining sample conditions at 2Hz. We learn embeddings for the unconditioned frames and integrate these embeddings into the conditions to match the frame rate of the videos. The combination of diverse frame-rate training schemes with the BiMOT module provides enhanced sparse condition control capability.

\noindent \textbf{Drop conditions} To enhance the robustness of the training process, the key views are trained in random order. There is a 50\% probability of dropping out the neighbor camera videos and an 20\% probability of dropping out the conditions. Our model does not require any post-refine process. It is capable of learning multi-view and temporal consistency simultaneously, which will enhance overall quality and realism.

\subsection{Inference Stage.}
\begin{figure}[h]
\centering
\includegraphics[width=1.0\linewidth]{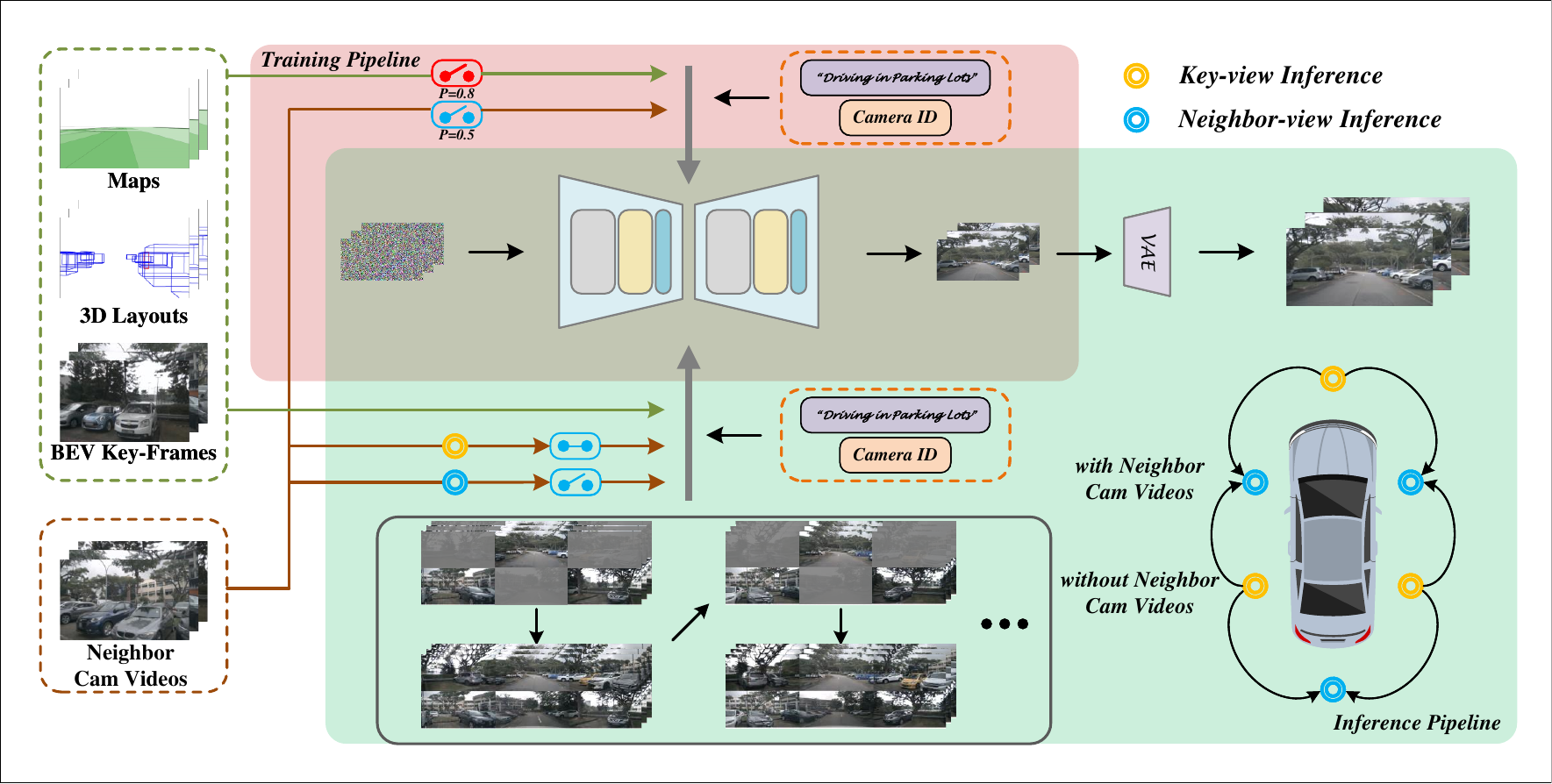}
\caption{The DriveScape pipeline operates on the LDM pipeline to generate street-view videos conditioned on scene annotations, BEV maps, and 3D bounding boxes for each view. Additionally, we also introduce keyframes conditions collaborating with temporal and spatial self-attention modules to achieve consistency in spatial and temporal dimensions. Our approach establishes a unified model for multi-view video generation without requiring complex post-processing or any post-refinement procedures. Furthermore, the Bi-Directional Modulated Transformer(BiMoT) module which comprises two cross-attention layers with opposite directions and one temporal self-attention layer enables effective alignment and synergy between various 3D road structural information to achieve precise control over video generation.}
\label{fig:network2}
\end{figure}

During the inference stage, our pipeline allows for an efficient generation of multi-view videos with consistency. Same as training stage, this pipeline consists of two patterns, key-view inference and neighbor-view inference as shown in Fig~\ref{fig:network2}. The key-view camera videos can be generated simultaneously without cross-view condition control efficiently, because the key-view camera positions are intentionally non-adjacent to each other. For instance, we have one camera at the forefront, while two others are stationed at the rear—one to the left and another to the right. This strategic placement ensures comprehensive coverage from multiple perspectives without direct overlap.

As a result of this configuration, the neighbour-view cameras, which consist of two at the front and one at the back, inherently acquire cross-view conditional information from key-view inference as their left and right immediate neighbours. This setup allows the neighbour-view inference stage to be consistent with neighbour-cam videos conditions, thereby fostering reasonable cross-view consistency.

Leveraging this arrangement, our inference pattern excels with its parallel processing capabilities. It requires only two consecutive iterations to generate multi-view videos, remarkably preserving cross-view consistency throughout. This efficient method not only speeds up the video production process but also ensures that the generated videos remain cross-view realistic.

\subsection{Bi-Directional Modulated Transformer}
\label{sec:BiDirectionalModulatedTransformer}
To integrate multi-conditions in different modalities into the UNet model and ensure that every condition is effectively articulated, we proposed a Bi-Directional Modulated Transformer(BiMoT) module. This module serves to align features from different modalities into the latent space, subsequently reinforcing the temporal coherence within the merged conditions, and finally injects the aligned conditions into the network. The specific structure is depicted in Fig. \ref{fig:network}, which comprises two cross-attention layers with opposite directions and one temporal self-attention layer. 

\begin{algorithm}
\caption{Flow Function}
\label{al:flow_func}
\begin{algorithmic}[1]
\Procedure{FlowFunction}{($F_{in}, F_{out}$)}  
        \State $Q = MLP(F_{in})$
        \State $K = MLP(F_{out})$
        \State $V = MLP(F_{out})$
        \State $O = \text{Softmax}\left( \frac{Q \cdot K}{\sqrt{d}} \right) \cdot V$
        \State $Out = \text{MLP}(O) \cdot \text{BN}(F_{out}) + \text{MLP}(O) + F_{out}$  
        \State \Return $Out$
        \State \textit{$d$ represents the number of channels of the value vectors.}  
\EndProcedure
\end{algorithmic}
\end{algorithm}

\noindent\textbf{Latent to Condition Attention} 
To address diverse conditions with different modalities, we employ an latent-to-condition cross-attention mechanism to standardize the feature representation. We utilize pixel-level features as a signal to merge various condition features. Specifically, as Algorithm 1 shown, we set $F_{in} = $ Condition features and $F_{out} = $ Latent features, we extract key $K$ and value features $V$ from the embedded condition data $\mathcal{E}(c)$ and obtain query vectors from image latents $\mathcal{L}(I)$. This cross-attention operation facilitates the flow of condition information into the image latent space, which serves as a natural unified representation for integrating all modalities.

\noindent\textbf{Temporal Attention.} We apply temporal self-attention to ensure that the merged features effectively propagate along the temporal dimension. Additionally, sparse condition information is incorporated into the image latent space, which naturally integrates all modalities into a unified representation. This allows temporal information from latent features to interact with sparse conditions over time, thereby guiding the generation process with sparse control.
This mechanism ensures consistent and sparse control over the temporal dimensions.

\noindent\textbf{Condition to Image Attention.} Following self-temporal attention, a condition-to-image cross-attention mechanism will be utilized. as Algorithm 1 shown, we set $F_{in} = $ Latent features and $F_{out} = $ condition features, this process facilitates the information of merged condition features to the image latents, ultimately enabling unified control over video generation.

\noindent\textbf{Coarse to Fine Condition Mixture.} BiMoT has the flexibility to accommodate various types of conditions, allowing for easy extension to additional conditions. We first introduce maps and 3D layouts, followed by BEV key-frames, and finally integrate neighbor camera videos. This approach enables generation control to progress from a coarse level driven by maps and 3D layouts to a finer level guided by pixel-level details, ensuring both structural coherence and detailed fidelity. This flexibility allows our model realize  fine-grained layout controllability with mono-condition and multi-condition unified control.

\section{Experiments}

In this section, we will evaluate our model based on several metrics. Firstly, we will provide a detailed description of our experimental setup, dataset, and evaluation criteria. Secondly, we will analyze the qualitative results of our methods. Finally, we will conduct several ablative studies, to prove the effectiveness of our method.

\subsection{Experiment Details}

\noindent\textbf{Dataset.} The training data is sourced from the real-world driving dataset nuScenes (\cite{nuscenes}), a prevalent dataset in BEV segmentation and 3D object detection for driving scenarios. We follow the official settings, utilizing 700 street-view scenes for training and 150 for validation. Our method considers 8 object classes and 8 road classes.

\noindent\textbf{Evaluation Metrics.}
To thoroughly evaluate the effectiveness of our approach in delivering realism, continuity and precise control over multiple conditions, we have selected four key metrics for comparison against existing multi-view image and video generation methods. For assessing realism, we rely on the widely recognized Fréchet Inception Distance (FID \cite{fid}) to measure the quality of our synthesized images. To ensure our videos exhibit consistent and fluid motion, we use the Fréchet Video Distance (FVD \cite{fvd}) as a measure of temporal coherence. Additionally, we estimate the levels of controllability with two perception-based evaluations. The CVT (\cite{cvt}) is for conducting BEV segmentation, and BEVFusion (\cite{bevfusion}) is a method for 3D object detection. Both of these assessment methods have been previously trained on the nuScenes dataset, providing a reliable standard for our performance experiments.

\noindent\textbf{Implement Details.} We implement our approach based on the official codebase of Diffusers (\cite{diffusers}), and a pre-trained SVD (\cite{svd}) video generation model. As mentioned before, we kept the spatial components of the pre-trained model frozen and focused on training the temporal aspects along with our condition guidance module. For training and validation, we compiled images into numerous sequences of 8-frame videos, each of which is resized into a size of 576 $\times$ 1024. All models are trained with the 8-bit AdamW optimizer (\cite{adamw8bit}) with a base learning rate of 1e-5 and a batch size of 8. The training process is executed on 8 NVIDIA 80G-A100 GPUs and was completed within 120 hours. To preserve the capabilities of the pre-trained model, we employ zero initialization (\cite{controlnet}) for the last layer of newly added modules and set the learning rate for these modules to 10 times the base learning rate. Additionally, we observed that using a temporal increasing classifier-free guidance scale from SVD leads to gradual distortion during iterative inference. Therefore, in all experiments, we fixed the scale at 2.5.

\subsection{Experiments Results}

\begin{figure*}[t]
\centering
\includegraphics[width=0.9\linewidth]{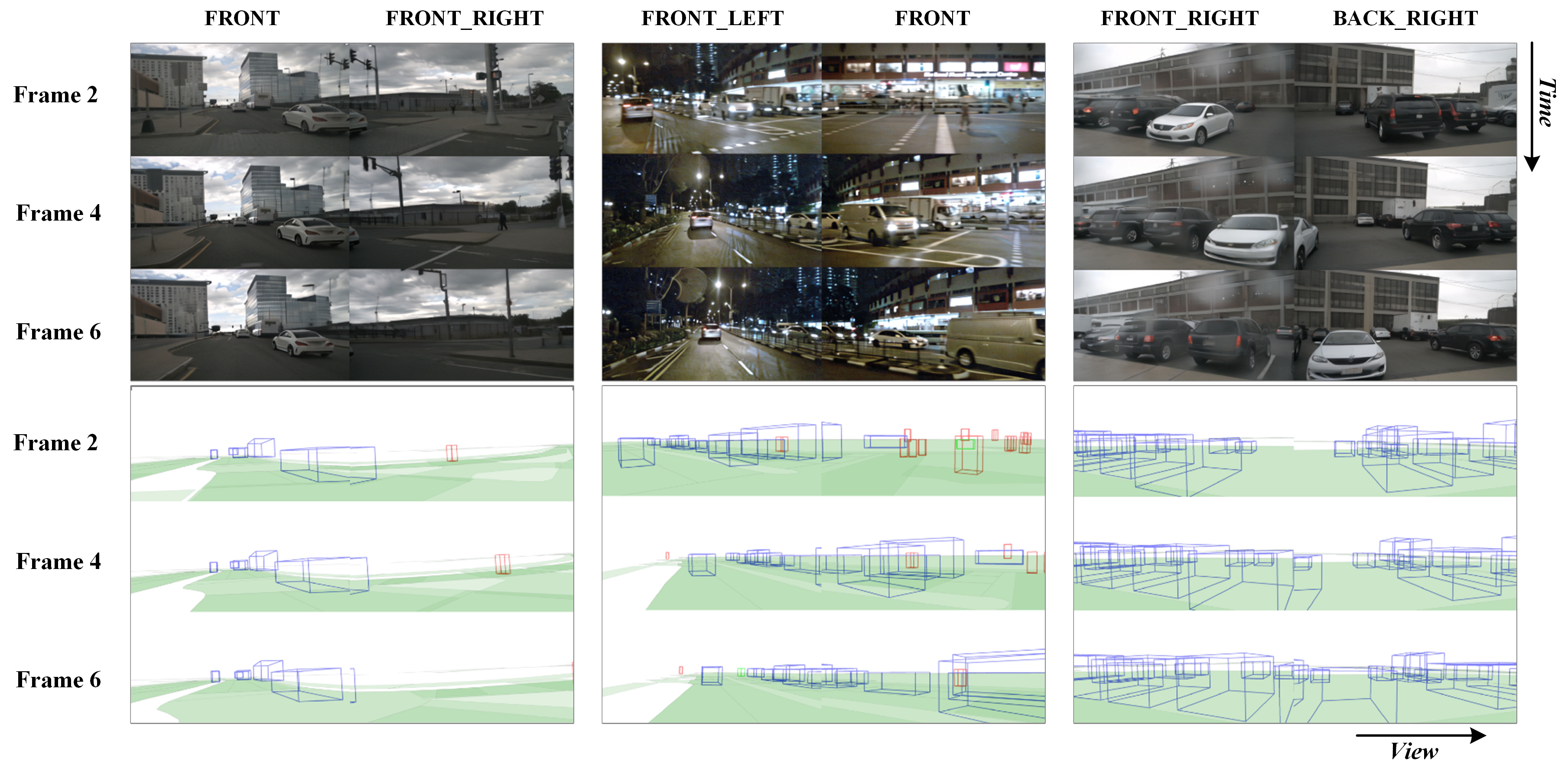}
\caption{Showcases for multi-view video generation of DriveScape; Our model achieves consistency across frames and viewpoints. Specifically, the vehicles generated by our model are accurately positioned following the provided 3D layouts, and the depiction of the streets closely aligns with the supplied projected maps. Importantly, even in cases where the provided bounding boxes extend beyond the visible area, our model showcases the capability to adhere to the specified conditions and generate the remaining objects with high fidelity.}
\label{fig:exp_showcase}
\vspace{-0.3cm}
\end{figure*}

\noindent\textbf{Qualitative Results.} We showcase the quality and consistency of our video generation outcomes in Fig. \ref{fig:exp_showcase}. Upon examination, it becomes evident that our model attains consistency across both frames and viewpoints. Specifically, vehicles generated by our model are accurately placed in accordance with the provided 3D layouts, and the delineation of the streets closely matches the supplied projected maps. Notably, even in instances where the provided bounding boxes extend beyond the visible area, our model demonstrates the ability to adhere to the specified conditions and generate the rest of the objects with high fidelity. 

\begin{figure}[ht]
\centering
\includegraphics[width=0.85\linewidth]{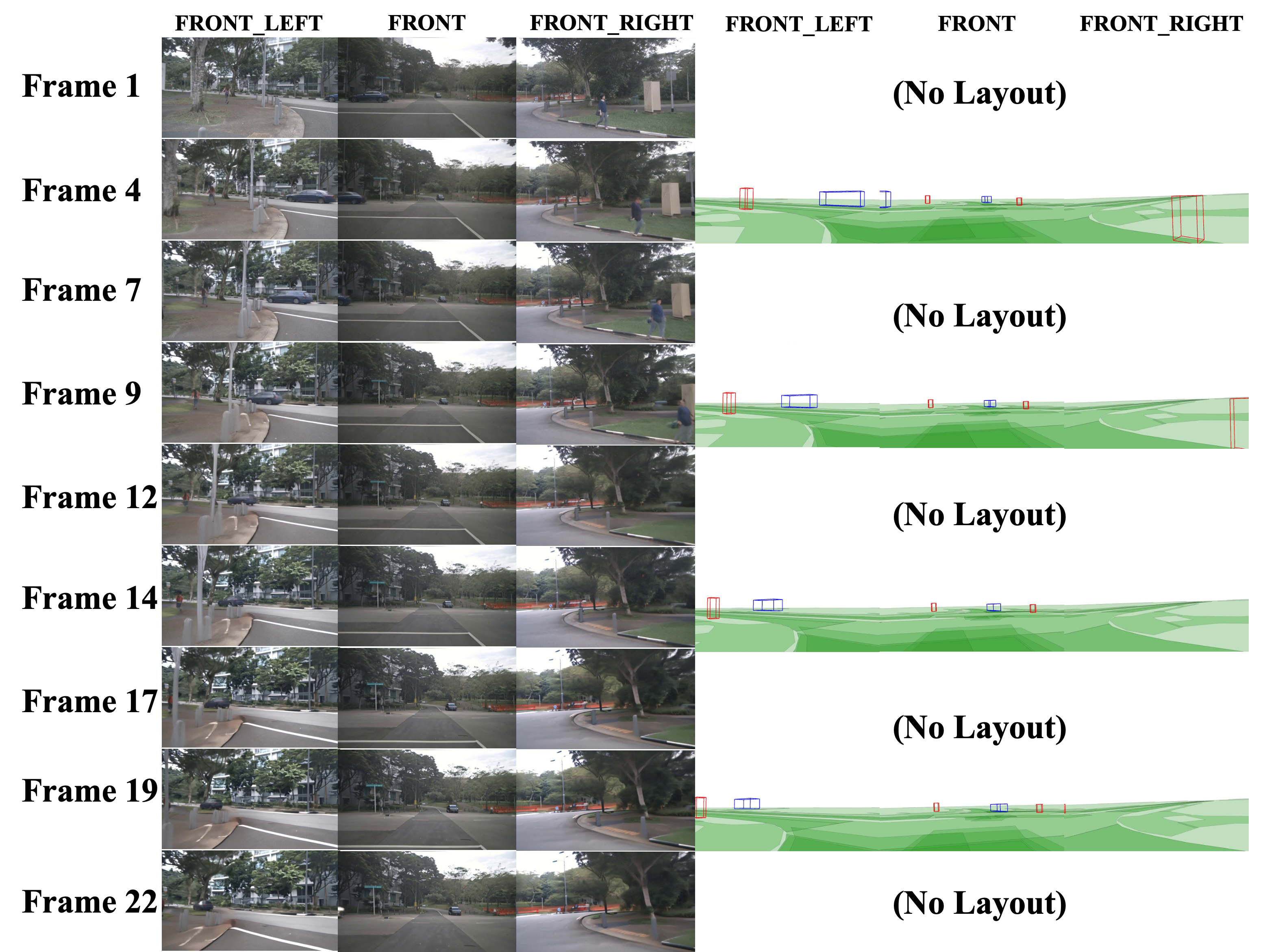}
\caption{Sparse Condition Control; Our model can generate consistent videos both in time and space with sparse condition with high resolution 576 $\times$ 1024 at 10 fps;}
\label{fig:zhutu}
\vspace{-0.3cm}
\end{figure}

\noindent\textbf{Sparse Condition Control} As shown in Fig~\ref{fig:zhutu}, our model can generate consistent videos both in time and space with sparse condition with high resolution 576 $\times$ 1024 and 10 fps;

\begin{table*}[htb]
  \centering
\begin{tabular}{l|c|c|c|c|c}
\hline
{Method} & {FID$\downarrow$} & {FVD$\downarrow$} & Road mIoU $\uparrow$ & Vehicle mIoU $\uparrow$ & Object NDS $\uparrow$ \\
\hline
Baseline
    & -
    & -
    & 73.68
    & 34.84
    & 41.21 \\
\hline
BEVGen
    & 25.54
    & -
    & 50.20 (-31.87\%)
    & 5.89 (-83.09\%)
    & - \\
BEVControl
    & 24.85
    & -
    & 60.80 (-17.48)
    & 26.80 (-23.08)
    & - \\ 
MagicDrive
    & 16.20
    & -
    & 61.05 (-12.63)
    & 27.01 (-7.83)
    & 23.32 (-17.89) \\
\hline
DriveDreamer
    & 52.60
    & 452.0
    & -
    & -
    & -   \\
DrivingDiffusion
    & 15.83
    & 332.0
    & 63.20 (-14.22)
    & -
    & 31.60 (-9.30)  \\
Drive-WM
    & 15.80
    & 122.70
    & 65.07 (-31.87)
    & -
    & 27.19 (-21.96)   \\
Delphi
    & 15.08
    & 113.5
    & -
    & -
    & 36.58 (-4.63)   \\
\hline
Ours
    & \textbf{8.34}
    & \textbf{76.39}
    & 64.43 (-9.25)
    & 28.86 (-5.98)
    & 36.5 (-4.71) \\
\hline
\end{tabular}
  \caption{The comparative analysis of our method against others using the nuScenes validation dataset is detailed for each task. We ensured that the experiments underwent similar settings compared to the other methods. Our end-to-end method outperformed most baseline models across the board without any post-refine process. The top-performing results have been highlighted in \textbf{bold} for clarity and emphasis. The parentheses indicate results of video generation work evaluated based on image generation.}
  \label{tab:comparisons}
\vspace{-0.3cm}
\end{table*}
\noindent\textbf{Quantitative Results.} 
We report quantitative experimental metrics on the nuScenes validation set, as shown in Table \ref{tab:comparisons}. Specifically, we utilize FID and FVD to assess the quality of generated images and videos, respectively. Following the approaches adopted by MagicDrive (\cite{magicdrive}) and DrivingDiffusion (\cite{drivediffusion}), we generate data using the validation set's labels as conditions. We then evaluate the generated data using a pre-trained perception model, calculating the decline in Road mIoU and Vehicle mIoU metrics compared to the original data. A lesser decline indicates that the generated data is more conducive to BEV perception models, demonstrating our framework's effectiveness in producing high-quality, perception-friendly synthetic data.

To align with previous studies, we resized the generated images to $224\times400$. We compared our results with three image generation models: BEVGen (\cite{bevgen}), BEVControl (\cite{bevcontrol}), and MagicDrive (\cite{magicdrive}), focusing on the image generation quality metric, FID. Our results significantly outperformed these models. It's noteworthy that video generation models typically have higher FID scores compared to image models. For example, Drive-WM using only an image model showed a better FID than the complete model, underscoring the effectiveness of our approach. We then assessed video generation quality, primarily using the FVD metric, and found substantial improvements over recent works like DriveDreamer (\cite{drivedreamer}), DrivingDiffusion (\cite{drivediffusion}), Drive-WM (\cite{intofuture}). In terms of BEV segmentation metrics, our method achieved better perceptual performance across the board, except for a slightly lower Road mIoU compared to DrivingDiffusion, which specifically trained a post-processing network to refine the main network's output. To maintain the simplicity of our end-to-end system, we performed no post-processing other than image resizing for perceptual evaluation. The perception models we used were based on single-image inference and did not leverage the temporal consistency of our generation results. Given our significant lead in the FVD metric, our algorithm holds great potential for training video BEV perception models.

    


\begin{figure}[t]
\centering
\includegraphics[width=1.0\linewidth]{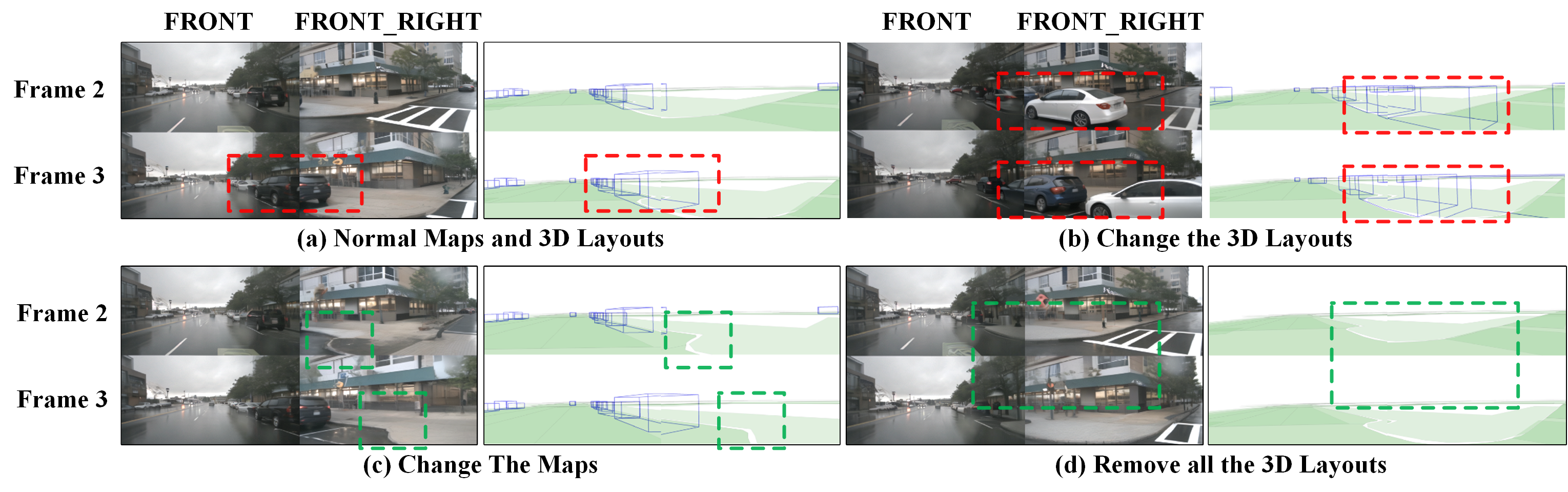}
\caption{Showcase for maps and 3D layouts control under different conditons. (a): control by the reference maps and 3D Layouts; (b) change 3D Layouts with more vehicles; (c) change the maps with curved pedestrian crosswalk; (d) remove all the 3D Layouts.}
\label{fig:exp_layoutcontrol}
\end{figure}

\noindent\subsection{Ablation Studies}
To refine and assess the efficacy of our proposed framework, we conducted a series of ablation studies, as shown in Tab. \ref{tab:ablations}. Initially, we adopt various strategies for implementing conditional guidance. If we remove the entire BiMoT module, the FID increases from 8.34 to 20.98, and the FVD rises from 76.39 to 142.81, indicating that this module is crucial for multi-condition encoding. Similarly, removing the temporal attention layer within BiMoT results in a significant performance drop, with FID and FVD increasing to 20.67 and 123.03, respectively. This demonstrates the importance of temporal modeling of conditions in BiMoT. Furthermore, during the testing phase, we separately removed the key frame condition and the neighbor condition, both of which led to a certain degree of performance decline, suggesting that these conditions are important for information exchange between views.

\begin{table}[ht]
\centering
\begin{tabular}{cccc|c|c}
\hline
temporal attn in BiMOT & BiMOT & key-frame cond & neighbor cond &          FID & FVD     \\
\hline
& $\checkmark$ & $\checkmark$ & $\checkmark$   &20.98     & 142.81  \\
& & $\checkmark$ & $\checkmark$       &20.67           & 123.03     \\
$\checkmark$ & $\checkmark$ &  & $\checkmark$ & 11.60      & 108.80   \\
$\checkmark$ & $\checkmark$ & $\checkmark$ & &12.18  &  111.36       \\
\hline
$\checkmark$ & $\checkmark$ & $\checkmark$ & $\checkmark$ &\textbf{{8.34}} & \textbf{76.39}  \\ 
\hline
\end{tabular}
\caption{Ablation Study. } 
\label{abl}
\label{tab:ablations}
\end{table}

\begin{figure}[ht]
\centering
\includegraphics[width=1.0\linewidth]{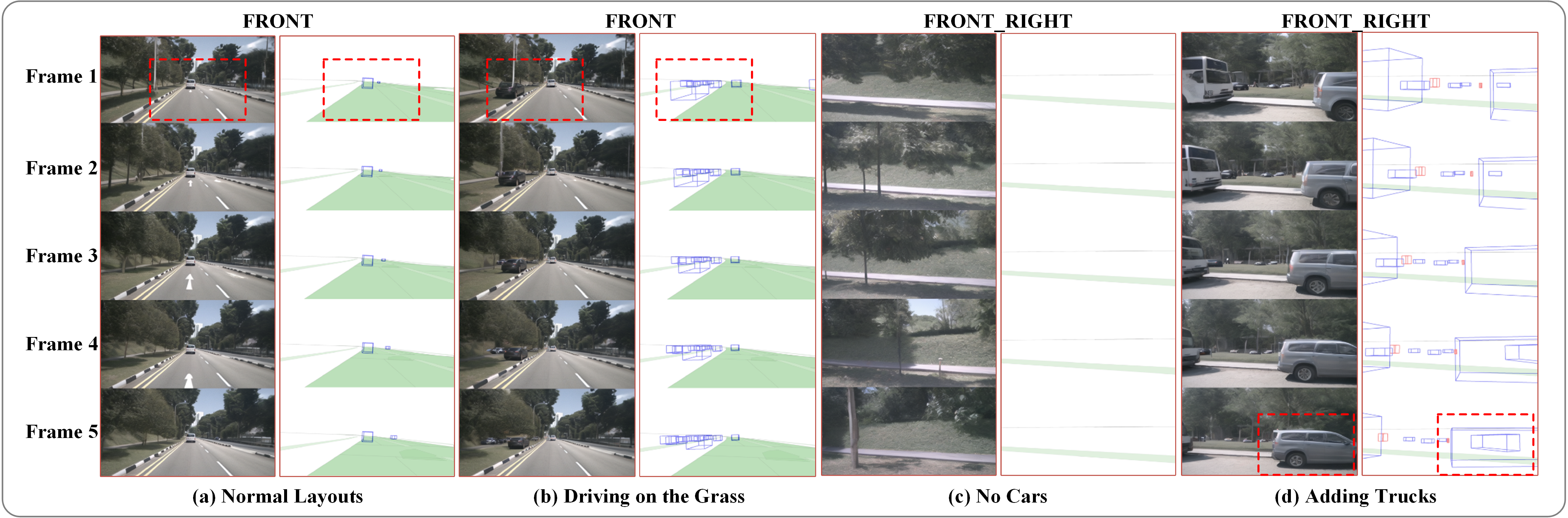}
\caption{Showcases for 3D layouts control under different conditions. DriveScape can precisely control dynamic objects. (a) reference layouts; (b) move car to the grass; (c) remove all cars; (d) add trucks to the scene.}
\label{fig:ablation_layout}
\end{figure}

\begin{figure}[ht]
\centering
\includegraphics[width=1.0\linewidth]{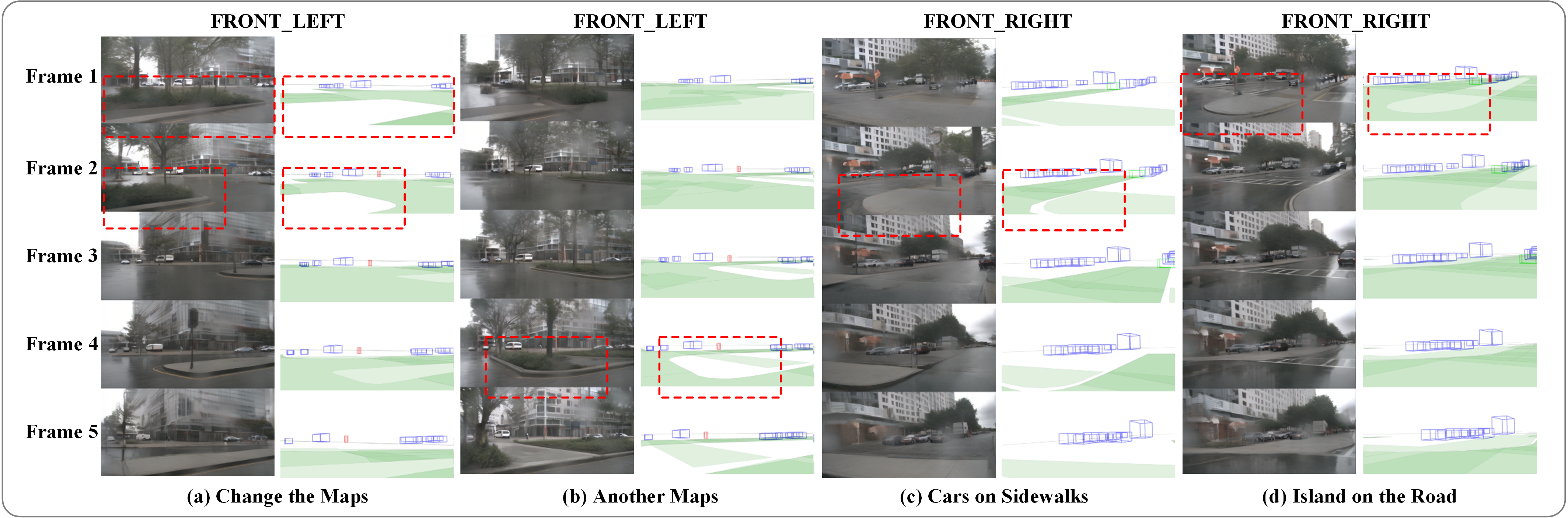}
\caption{Showcases for maps control under different conditions. DriveScape can precisely control static background. (a) replace driable area with green belts; (b) replace the driable area with green belts; (c) replace the driable area with sidewalks; (d) add island on the road;   }
\label{fig:ablation_maps}
\end{figure}

\subsection{Visualization of Difference Settings.}

In our visual showcases, we demonstrate the model's proficiency in manipulating video predictions of a single scene under a range of control conditions. The control conditions employed in these experiments were sourced from frames of identical perspectives, randomly chosen from other scenes. This methodology enabled us to conduct experiments on the decoupling of control conditions, testing the robustness of our approach across varied and unrelated scenarios.

As depicted in Fig. \ref{fig:ablation_layout}, the illustration showcases results governed by different 3D layout controls. Utilizing a single reference frame captured from a segment on the highway, we first illustrate a typical example of vehicles driving normally. Subsequently, we transition the scene to depict the same vehicle traversing a grassy terrain. Remarkably, even in this unconventional scenario, our model can also maintains temporal consistency. Further exploration involved either removing or adding large vehicles to the scene. The results, as can be seen, are also aligned with the specified conditions, demonstrating the precise control our model exerts over the generated content.

Additionally, As depicted in Fig. \ref{fig:ablation_maps}, we explore the influence of different maps on the video generation process. The scenes generated are impressively congruent with the shapes of the given map regions. Moreover, the consistency of layout features is not compromised by changes in the map, proving the control effectiveness of our model. This series of experiments serves as strong evidence of our model, confirming its ability to reliably render the specified control conditions into precise and coherent visual narratives in the generated video content.

\section{Conclusion and Future Works}
\textbf{Conclusion.} The paper introduces DriveScape as the first end-to-end multi-view driving video generative model. We have developed a unified model for consistent multi-view video generation without the need for complex post-processing or any post-refinement procedures. This end-to-end framework enables streamlined, effective, and efficient multi-view condition-guided video generation tasks. Additionally, we have introduced a Bi-Directional Modulated Transformer module to facilitate diverse forms of condition guidance in video generation tasks. Our demonstrations show that we have achieved decoupled and precise control of dynamic foreground and static background. Furthermore, our approach has achieved state-of-the-art video synthesis performance on the nuScenes dataset, excelling not only in the generation quality of multi-view images and videos but also in various perception applications. In summary, our method is superior to others in terms of generation quality and ease of application.

\textbf{Future Works.} Our DriveScape can be trained with higher spatial-temporal resolution to achieve better results. Similarly to DrivingDiffusion (\cite{drivediffusion}), we can also train a post-processing network, further narrowing the distribution gap between generated and real data. Additionally, designing appropriate training strategies to mitigate the memory overhead caused by multi-view inputs is crucial. Lastly, This approach will enable the incorporation of more multi-view constraint modules during the training phase, which is also a potential improvement.

\bibliography{iclr2025_conference}
\bibliographystyle{iclr2025_conference}


\end{document}